\pdfoutput=1

\documentclass[11pt]{article}

\usepackage[preprint]{acl}

\usepackage{times}
\usepackage{latexsym}
\usepackage{graphicx}
\usepackage{array}
\usepackage{tabularx}
\usepackage{float}
\usepackage{tabularx,colortbl}
\usepackage{multirow}
\usepackage{comment}

\usepackage[T1]{fontenc}

\usepackage[utf8]{inputenc}

\usepackage{microtype}

\usepackage{inconsolata}

\title{Abdelhak at SemEval-2024 Task 9 : Decoding Brainteasers, The Efficacy of Dedicated Models Versus ChatGPT}

\author{Abdelhak Kelious \\
  ATILF, University of Lorraine\\
  And CNRS / France \\
  \texttt{abdelhak.kelious@univ-lorraine.fr} \\\And
  Mounir Okirim \\
  ESIEA, Graduate School
  \\of Engineering / France
 \\
  \texttt{okirim@et.esiea.fr} \\}

\begin{document}
\maketitle
\begin{abstract}
This study introduces a dedicated model aimed at solving the BRAINTEASER task 9 \cite{jiang2023brainteaser}, a novel challenge designed to assess models' lateral thinking capabilities through sentence and word puzzles. Our model demonstrates remarkable efficacy, securing Rank 1 in sentence puzzle solving during the test phase with an overall score of 0.98. Additionally, we explore the comparative performance of ChatGPT, specifically analyzing how variations in temperature settings affect its ability to engage in lateral thinking and problem-solving. Our findings indicate a notable performance disparity between the dedicated model and ChatGPT, underscoring the potential of specialized approaches in enhancing creative reasoning in AI.
\end{abstract}

\section{Introduction}
The BRAINTEASER task \cite{jiang2023brainteaser} aims to challenge the lateral thinking abilities of models, setting it apart from traditional tasks focused on vertical logical reasoning. It introduces lateral thinking puzzles in the form of multiple-choice questions to test the models' ability to think creatively and challenge common sense associations. The goal is to identify the gap between human and model performances in creative thinking, highlighting the need for progress in AI's creative reasoning abilities. NLP (Natural Language Processing) transformer models have revolutionized text understanding and generation with their architecture capable of processing word sequences more efficiently. For multiple-choice questions, these models utilize their ability to understand context and language nuances to select the most appropriate answer from several options. Thanks to deep learning and attention mechanisms, they excel in various NLP tasks, significantly improving the accuracy and relevance of responses generated in complex contexts.
The integration of NLP transformer models into the BRAINTEASER task aims to explore their ability to solve lateral thinking puzzles in the form of multiple-choice questions. This approach highlights the challenges posed by deep language understanding and the creativity required to surpass traditional logical reasoning. It emphasizes the importance of advancing in the development of models capable of navigating beyond common sense associations, encouraging innovation in the interpretation and generation of complex and nuanced responses.
In our study, we will explore the ability of language models to handle this task, with the following main contributions of this paper :

\begin{itemize}
\item Development of a dedicated model for this task with a good result for the sentence puzzle task (Rank 1 in the test phase).
\item A comparative analysis with ChatGPT: Specifically, the relationship of temperature with lateral thinking and performance.

\end{itemize}

\section{Shared Task Description}
The BRAINTEASER Shared Task 9 \cite{jiang2023brainteaser} is a Question Answering (QA) task based on evaluating the capacity of language models to engage in lateral thinking and to solve puzzles that require unconventional thinking. BRAINTEASER comprises two distinct subtasks: Sentence Puzzle and Word Puzzle, both of which involve defying commonsense "defaults" but through different methodologies.
\begin{itemize}
\item Sentence Puzzle: Create sentence-based brain teasers where the challenge lies in interpreting sentence snippets in a way that goes against commonsense expectations.

\item Word Puzzle: Design word-based brain teasers that require rethinking the default meanings of words, with a focus on the composition of letters in the target question.
\end{itemize}
Both tasks include an adversarial subset, created by manually modifying the original brain teasers without changing their latent reasoning path. They construct adversarial versions of the original data in two ways:
\begin{itemize}
    \item (SR) Semantic Reconstruction  rephrases the original question without changing the correct answer and the distractors.
    \item (CR) Context Reconstruction  keeps the original reasoning path but changes both the question and the answer to describe a new situational context
\end{itemize}
Distractors are generated by identifying the implicit and explicit premises of a puzzle and then manually overwriting these premises, ensuring they remain incorrect but challenging.

The  BRAINTEASER \cite{jiang2023brainteaser} paper reveals a significant gap between human performances and AI models, and underscores the need to enhance lateral reasoning in language models.

\section{Related Work}

The task of commonsense reasoning has long been a challenge for deep learning and has been the subject of research for several years, accompanied by various benchmarks such as \cite{nie-etal-2020-adversarial}, which introduces a new large-scale NLI benchmark dataset created through an adversarial process involving humans and models. This improves NLI models' performance on popular benchmarks and reveals their weaknesses, offering a dynamic framework for continuous improvement in natural language understanding. A study demonstrated a simple and unsupervised method for commonsense reasoning using language models trained on vast text corpora, significantly outperforming state-of-the-art methods on Pronoun Disambiguation Problems and the Winograd Schema Challenge without the need for annotated knowledge bases or manually engineered features \cite{trinh2019simple}.

Transformer models like BERT \cite{devlin2019bert}, GPT \cite{brown2020language}, and their variants have revolutionized natural language understanding, including question answering \cite{qu2019bert}. Their architecture captures semantic and contextual nuances \cite{ethayarajh2019contextual} \cite{zhang2020semantics}, proving exceptionally effective in comprehending and responding to complex inquiries. By training on extensive text corpora, they develop a deep understanding, enabling them to identify the most plausible answers among multiple choices \cite{roy2023analysis} \cite{ravi2023vlc}.

 Large pretrained language models (PLMs) can achieve near-human performance on commonsense reasoning tasks by generating contrastive explanations that highlight the key attributes needed to justify correct answers. This approach not only improves performance on commonsense reasoning benchmarks but also produces explanations judged by humans as more relevant and understandable \cite{paranjape-etal-2021-prompting}

Recent studies reveal that ChatGPT has notable capabilities to effectively solve a variety of problems in several languages, including the task of answering questions. Moreover, its performance improves with each new version. ChatGPT excels in certain areas but also has its limitations in terms of consistency and complex reasoning tasks.\cite{tan2023evaluation}.

\section{Proposed Approach } 
\subsection{Methodology}
In our study, we have developed a model based on transformers for multiple-choice questions, where each option is combined with the question to form separate pairs. These pairs are then pre-processed as distinct inputs for the already pre-trained model. The preprocessing includes adding special tokens like [CLS] at the beginning and [SEP] to separate the question from the choice. Each pre-processed question-choice pair is passed through the transformer model, which encodes each pair using its bidirectional attention mechanism, allowing every word in the pair to capture the context of the entire sentence and the related choice. For each question-choice pair, the model generates a feature vector from the output associated with the [CLS] token, which serves as a summary of the information contained in the pair. This means that for a question with four answer choices, the model would be run four times (once for each question-choice pair). This process allows for the consideration of the full context of the question as well as that of each individual answer choice, which is crucial for understanding which choice best answers the question. 
\begin{figure*}[h]
    \centering
    \includegraphics[width=0.94\textwidth]{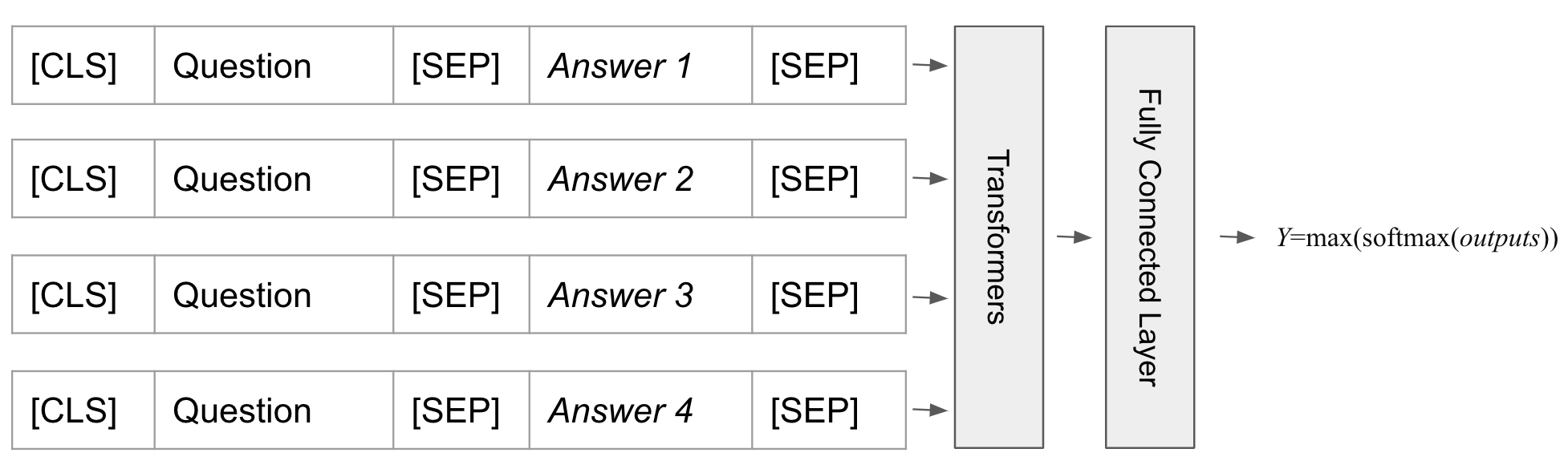}
    \caption{The overall architecture for predicting BRAINTEASER}
    \label{fig:model_arch}
\end{figure*}
\begin{figure*}[h]
    \centering
    \includegraphics[width=0.94\textwidth]{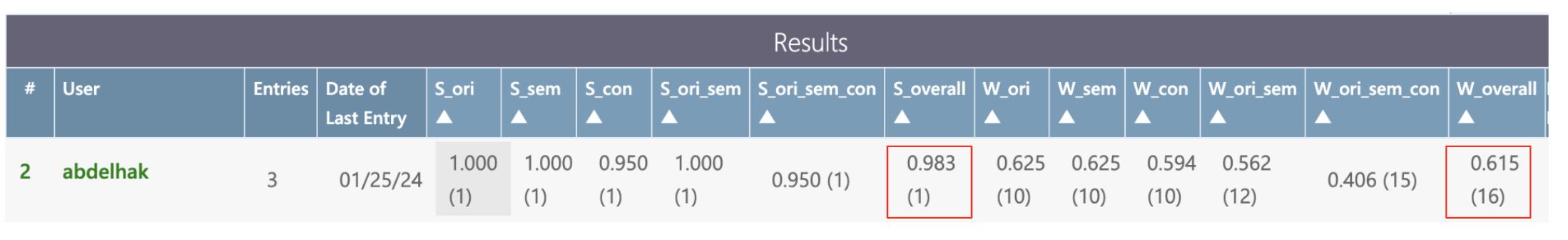}
    \caption{The Ranking Leaderboard Displaying Our Position}
    \label{fig:leaderboard}
\end{figure*}
The feature vector for each question-choice pair is then passed through a dense (or fully connected) layer, which reduces the vector's dimensionality to a number corresponding to the number of classes or answer categories. After the dense layer, a softmax activation function is applied to convert the scores into probabilities.

The softmax function is ideal for classification tasks because it transforms the scores into a set of probabilities that sum up to 1, making the scores directly interpretable as the probabilities that each choice is the correct answer. Figure \ref{fig:model_arch} illustrates the prediction process described above.

The prediction formula can be expressed as follows in our model:

Each question-choice pair \((Q, C_i)\) is pre-processed to form an input sequence \(X_i\) by concatenating the question \(Q\) with each choice \(C_i\) and adding special tokens:
   \[X_i = [CLS] + Q + [SEP] + C_i + [SEP]\]
The transformer model processes each \(X_i\) separately to encode the pair, utilizing its bidirectional attention mechanism. The output for each token in \(X_i\) is obtained, but we are specifically interested in the output associated with the [CLS] token, \(T_{[CLS],i}\), which captures the contextualized representation of the pair:
   \[T_{[CLS],i} = TransformerModel(X_i)\]
The feature vector \(F_i\) is extracted from the transformer output associated with the [CLS] token for each question-choice pair:
   \[F_i = ExtractFeatureVector(T_{[CLS],i})\]
Each feature vector \(F_i\) is passed through a dense layer to reduce its dimensionality to the number of classes \(N\), resulting in a reduced feature vector \(R_i\):
   \[R_i = DenseLayer(F_i)\]
The softmax activation function is applied to \(R_i\) to convert the scores into probabilities \(P_i\), indicating the likelihood that each choice is the correct answer:
   \[P_i = Softmax(R_i) = \frac{e^{R_{i}}}{\sum_{j=1}^{N} e^{R_{j}}}\]

Where:
\begin{itemize}

    \item \(Q\) represents the question.
    \item \(C_i\) represents the \(i\)th answer choice.
    \item \(X_i\) is the input sequence formed by concatenating \(Q\) and \(C_i\) with special tokens.
    \item \(T_{[CLS],i}\) is the transformer output for the [CLS] token for the \(i\)th question-choice pair.
    \item \(F_i\) is the feature vector extracted from \(T_{[CLS],i}\).
    \item \(R_i\) is the reduced feature vector after passing \(F_i\) through a dense layer.
    \item \(P_i\) represents the probabilities that each choice \(C_i\) is the correct answer, obtained after applying the softmax function to \(R_i\).
\end{itemize}

\subsection{Evaluation Method}

The BRAINTEASER task proposes the following evaluation system, each system is evaluated based on the following two accuracy metrics:

\textbf{Instance-based Accuracy}: They consider each question (original/adversarial) as a separate instance. They  report accuracy for the original and its adversaries.

\textbf{Group-based Accuracy}: Each question and its associated adversarial instances form a group, and a system will only receive a score of 1 when it correctly solves all questions in the group. 

The final score corresponds to the average of all the scores.
\subsection{Results}
We trained our model using the pre-trained language model DeBERTa-v3-base \cite{he2023debertav3} over 5 learning epochs, with a learning rate of 5e-5 and a batch size of 16. The results obtained are presented in official Leaderboard of the task in the evaluation phase \ref{fig:leaderboard}.

Our model stands out for its good performance in sentence-type puzzles, ranking first with with an average accuracy score of \textbf{0.98} ( leaderboard \ref{fig:leaderboard}) . This means it excels particularly in thinking challenges where the puzzle, often contrary to common sense, is based on sentence excerpts. On the other hand, for word-based puzzles, which require finding a solution that goes against the usual meaning of words by focusing on the letter composition of the posed question, our model shows lower performance. It ranks 16th with a total score of \textbf{0.61} . This performance difference suggests that, although our model is very skilled at solving puzzles involving the understanding and manipulation of sentences, it could benefit from improvement in the area of word-based puzzles. This indicates an opportunity to deepen our research and development efforts on word-type puzzles to enhance the versatility and overall effectiveness of our model.

\section{ChatGPT Analysis }
\subsection{Zero-shot Predictions}
Given that we are currently in the era of ChatGPT, it's challenging to approach our study without including a comparison to evaluate the role of this task in relation to ChatGPT. We crafted a simple and explicit prompt with ChatGPT turbo  3.5 on February 5, 2024, assessing ChatGPT's logical reasoning ability using various prompts in a qualitative manner. However, we faced challenges in determining the optimal prompt, as the same input does not always lead to the desired output. Hallucinations related to conversation history were resolved by initiating a new session for each iteration. In the end, we settled on the following prompt:
\begin{quote}
“”” 

Question ?

liste of choices : 

	1- Answer 1.
 
	2- Answer 2.
 
	3- Answer 3.
 
	4- Answer 4. 
 
Response should be in json format : 

\{ “answer”: Number of the choice \}

“””

\end{quote}

We achieved a total score of \textbf{0.59} for the sentence-puzzle task and \textbf{0.27} for the word-puzzle task, scores that do not necessarily match the expected performance for a  model like ChatGPT. This suggests that, although ChatGPT was not specifically trained for this task, it might not be able to compete with models that were specially designed for it. ChatGPT was trained on a vast dataset, but it is assumed that most of this data is well-structured and more aligned with linear thinking rather than lateral thinking, which explains its moderate performance in this area.

\subsection{The Effect of Temperature}
The temperature parameter in language models for natural language processing is a hyperparameter used to control the diversity of predictions made by the model during text generation. Temperature adjusts the likelihood of predictions based on their calculated probability, thereby influencing the level of risk or surprise in the choice of generated words. Adjusting the temperature allows for control over the trade-off between creativity and safety in text generation. Finding the right temperature depends on the specific application, the domain of use, and preferences for the balance between innovation and reliability in the generated responses. A low temperature close to 0 produces more conservative and repetitive responses, while a high temperature close to 1 yields more varied and creative responses.

\textbf{Is there a relationship between temperature and lateral thinking ?} Although the temperature setting in language models and lateral thinking operate in different domains, they share a common goal of fostering creativity and innovation by breaking conventions and exploring possibilities beyond those that are immediately obvious. Lateral thinking encourages questioning assumptions and considering a variety of different perspectives. Similarly, by adjusting the temperature to favor less likely word selections, a language model can "think" more laterally, exploring linguistic options that would not be considered at a lower temperature. Therefore, we will measure the performance of ChatGPT based on temperature, relationship between temperature and lateral thinking.We will launch several runs by increasing the temperature from 0 to 1.2

\begin{figure}[htp]
    \centering
    \includegraphics[width=0.47\textwidth]{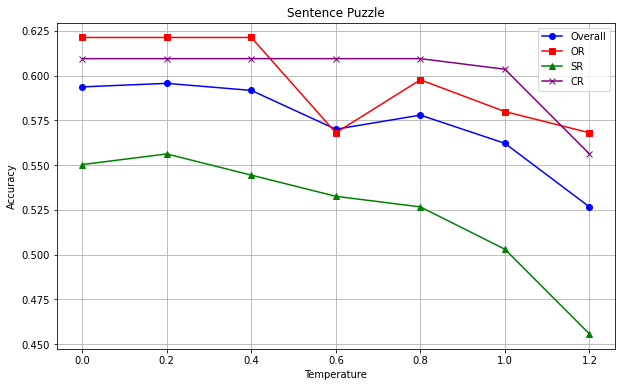}
    \caption{ChatGPT Performance Across Different Temperatures (Sentence puzzle) }
    \label{fig:SentencePuzzle}
\end{figure}
\begin{figure}[htp]
    \centering
    \includegraphics[width=0.47\textwidth]{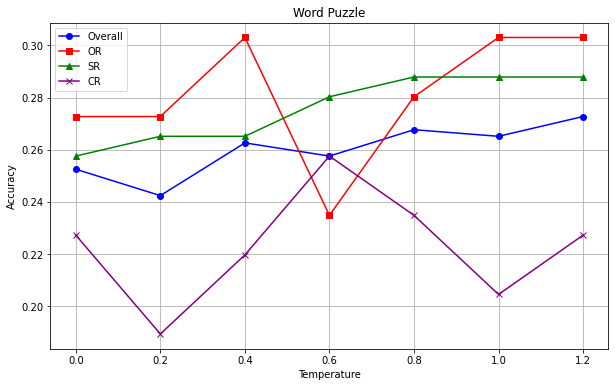}
    \caption{ChatGPT Performance Across Different Temperatures (Word puzzle) }
    \label{fig:word_puzzle}
\end{figure}

\textbf{Sentence Puzzle :} The graphic \ref{fig:SentencePuzzle} represents four data series corresponding to different test scenarios for the sentence puzzle task: Overall, OR (Original), SR (Semantic Reconstruction), and CR (Context Reconstruction). "Overall" indicates a benchmark or an overall average of performance, while OR shows stable results, suggesting a consistent baseline. CR follows a trend similar to OR, indicating that contextual reconstruction performs comparably to the original. In contrast, SR shows a notable degradation in performance towards the end, which could suggest that the semantic reconstruction method is less stable or effective under certain conditions. The data set suggests that while OR and CR methods maintain a degree of consistency, SR might involve a riskier or more innovative approach, which could be likened to a "higher temperature" in the context of lateral thinking, leading to more varied and potentially less predictable outcomes. However, increasing the temperature does not allow the model to perform better on a task, on the contrary, performance decreases.

\textbf{Word puzzle : }In the case of word puzzles \ref{fig:word_puzzle}, it is difficult to conclude as there are no clear trends observed. However, for the overall general case, it is noted that performance increases very slightly with temperature, which stands out in comparison to the sentence puzzle task, potentially because word puzzles better illustrate lateral thinking. In this case, the focus is not on the sentence, which contains more semantic aspects. The answer in this task violates the default meaning of the word and focuses on the letter composition of the target question.
\section{Conclusion}
Our research underscores the significance of dedicated models in advancing AI's capability to solve complex lateral thinking tasks, as exemplified by our model's top-ranking performance in the BRAINTEASER sentence puzzles. The comparative analysis with ChatGPT highlights the limitations of general-purpose models in specific creative reasoning challenges, despite their overall versatility. The study also reveals the nuanced role of temperature settings in modulating ChatGPT's performance, offering insights into optimizing AI models for enhanced creativity and lateral thinking. Future work should focus on bridging the gap in word puzzle performance and further refining the balance between creativity and logical reasoning in AI systems.

\bibliography{custom}

\end{document}